\documentclass[runningheads]{llncs}
\usepackage[sectionbib,numbers]{natbib}  
\usepackage{amsmath}
\usepackage{graphicx}
\usepackage[colorinlistoftodos]{todonotes}
\usepackage[colorlinks=true, allcolors=blue]{hyperref}

\usepackage{comment}
 
\usepackage{color}
\definecolor{Orange}{rgb}{1,0.5,0}
\definecolor{Red}{rgb}{1,0,0}
\definecolor{Blue}{rgb}{0,0,1}

\newcommand{\BL}[1]{\textsf{\textbf{\textcolor{Red}{\footnotesize [BL: #1]}}}}

\usepackage{bm}
\usepackage{amsfonts} 
\usepackage{amssymb} 
\usepackage{stmaryrd} 

\usepackage{booktabs}
\usepackage{longtable}
\usepackage{array}
\usepackage{multirow}
\usepackage[linesnumbered,ruled,vlined]{algorithm2e}

\usepackage{graphicx}
\usepackage{subfigure}

\setcitestyle{square} 

\usepackage[misc]{ifsym} 

\begin{document}
%

\title{Synthetic Oversampling of Multi-Label Data based on Local Label Distribution
}
%
%
\author{Bin Liu \Letter 
\and Grigorios Tsoumakas
}
\authorrunning{B. Liu and G. Tsoumakas}
%
\institute{School of Informatics \\ Aristotle University of Thessaloniki \\ Thessaloniki 54124, Greece \\ \email{\{binliu,greg\}@csd.auth.gr}}
\maketitle              
\begin{abstract}
Class-imbalance is an inherent characteristic of multi-label data which affects the prediction accuracy of most multi-label learning methods. 
One efficient strategy to deal with this problem is to employ resampling techniques before training the classifier. 
Existing multi-label sampling methods alleviate the (global) imbalance of multi-label datasets. 
However, performance degradation is mainly due to rare sub-concepts and overlapping of classes that could be analysed by looking at the local characteristics of the minority examples, rather than the imbalance of the whole dataset. 
We propose a new method for synthetic oversampling of multi-label data that focuses on local label distribution to generate more diverse and better labeled instances.  
Experimental results on 13 multi-label datasets demonstrate the effectiveness of the proposed approach in a variety of evaluation measures, particularly in the case of an ensemble of classifiers trained on repeated samples of the original data.


\end{abstract}
\section{Introduction}
In multi-label data, each example is typically associated with a small number of labels, much smaller than the total number of labels. This results in a sparse label matrix, where a small total number of positive class values is shared by a much larger number of example-label pairs. From the viewpoint of each separate label, this gives rise to class imbalance, which has been recently recognized as a key challenge in multi-label learning~\cite{Charte2015a,Charte2015,Daniels2017,Liu2018MakingImbalance,Zhang2015a}. 

Approaches for handling class imbalance in multi-label data can be divided into two categories: a) reducing the imbalance level of multi-label data via resampling techniques, including  synthetic data generation~\cite{Charte2014,Charte2015a,Charte2015,Charte2019}, and b) making multi-label learning methods resilient to class imbalance ~\cite{Daniels2017,Liu2018MakingImbalance,Zhang2015a}. This work focuses on the first category, whose approaches can be coupled with any  multi-label learning method and are therefore more flexible.


Existing resampling approaches for multi-label data focus on class imbalance at the global scale of the whole dataset. However, previous studies of class imbalance in binary and multi-class classification~\cite{napierala2016types,Saez2016} have found that the distribution of class values in the local neighbourhood of minority examples, rather than the global imbalance level, is the main reason for the difficulty of a classifier to recognize the minority class. We hypothesize that this finding is also true, and even more important to consider, in the more complex setting of multi-label data, where it has not been examined yet. 

\begin{figure}
\begin{center}
\includegraphics[width=1.0\textwidth]{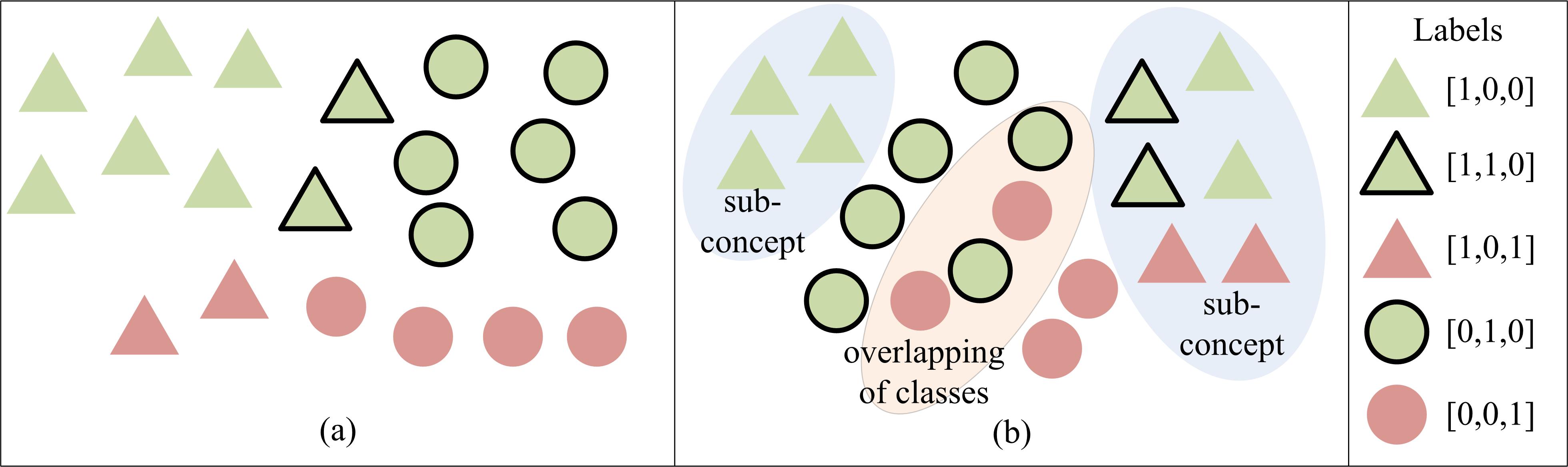}
\caption{Two 2-dimensional multi-label datasets (a) and (b) concerning points in a plane characterized by three labels. On the right we see the five different label combinations that exist in the datasets.}
\label{fig:ExampleDif}
\end{center}
\end{figure}

Consider for example the 2-dimensional multi-label datasets (a) and (b) in Fig.\ref{fig:ExampleDif} concerning points in a plane. The points are characterized by three labels, concerning the shape of the points (triangles, circles), the border of the points (solid, none) and the color of the points (green, red). These datasets have the same level of label imbalance. Yet (b) appears much more challenging due to the presence of sub-concepts for the triangles and the points without border and the overlap of the green and red points as well as the points with solid and no border.


This work proposes a novel multi-label synthetic oversampling method, named MLSOL, whose seed instance selection and synthetic instance generation processes depend on the local distribution of the labels. This allows MLSOL to create more diverse and better labelled synthetic instances. Furthermore, we consider the coupling of MLSOL and other resampling methods with a simple but flexible ensemble framework to further improve its performance and robustness. 
Experimental results on 13 multi-label datasets demonstrate the effectiveness of the proposed sampling approach, especially its ensemble version, for three different imbalance-aware evaluation metrics and six different multi-label methods.




The remainder of this paper is organized as follows. Section 2 offers a brief review of methods for addressing class imbalance in multi-label data. Then, our approach is introduced in Section 3. Section 4 presents and discusses the experimental results. 
Finally, Section 5 summarizes the main contributions of this work. 


\section{Related Work}

A first approach to dealing with class imbalance in the context of multi-label data is to utilize the resampling technique, which is applied in a pre-processing step and is independent of the particular multi-label learning algorithm that will be subsequently applied to the data. LP-RUS and LP-ROS are two twin sampling methods, of which the former removes instances assigned with the most frequent labelset (i.e. particular combination of label values) and the latter replicates instances whose labelset appears the fewest times~\cite{charte2013}. 

Instead of considering whole labelset, several sampling methods alleviate the imbalance of the dataset in the individual label aspect, i.e. increasing the frequency of minority labels and reducing the number of appearances of majority labels.  
ML-RUS and ML-ROS simply delete instances with majority labels and clone examples with minority labels, respectively~\cite{Charte2015}. 
MLeNN eliminates instances only with majority labels and similar labelset of its neighbors in a heuristic way based on the Edited Nearest Neighbor (ENN) rule~\cite{Charte2014}. 
To make a multi-label dataset more balanced, MLSMOTE randomly selects instance containing minority labels and its neighbors to generate synthetic instances which are associated with labels that appear more that half times of the seed instance and its neighbors according to $Ranking$ strategy~\cite{Charte2015a}. 

REMEDIAL tackles the concurrence of labels with different imbalance level in one instance, of which the level is assessed by $SCUMBLE$, by decomposing the sophisticated instance of into two simpler examples, but may introduce extra confusions into the learning task, i.e. there are several pairs of instances with same features and disparate labels~\cite{Charte2019}. 
The REMEDIAL could be either a standalone sampling method or the prior part of other sampling techniques, i.e. RHwRSMT combines REMEDIAL with MLSMOTE~\cite{Charte2019a}.



Apart from resampling methods, another group of approaches focuses on multi-label learning method handling the class-imbalance problem directly. 
Some methods deal with the imbalance issue of multi-label learning via transforming the multi-label dataset to several binary/multi-class classification problems. COCOA converts the original multi-label dataset to one binary dataset and several multi-class datasets for each label, and builds imbalance classifiers with the assistance of sampling for each dataset~\cite{Zhang2015a}. SOSHF transforms the multi-label learning task to an imbalanced single label classification assignment via cost-sensitive clustering, and the new task is addressed by oblique structured Hellinger decision trees~\cite{Daniels2017}.
Besides, many approaches aims to modify current multi-label learning methods to handle class-imbalance problem. ECCRU3 extends the ECC resilient to class imbalance by coupling undersampling and improving of the exploitation of majority examples\cite{Liu2018MakingImbalance}. Apart from ECCRU3, the modified models based on neural network~\cite{Tepvorachai2008,Li2013,Sozykin2017}, SVM~\cite{Cao2016}, hypernetwork~\cite{Sun2017} and BR~\cite{Chen2006,Dendamrongvit2010UndersamplingDomains,Tahir2012,Wan2017} have been proposed as well.
Furthermore, other strategies, such as representation learning \cite{LiWang16}, constrained submodular minimization \cite{Wu:2016} and balanced pseudo-label \cite{Zeng2014}, have been utilized to address the imbalance obstacle of multi-label learning as well.

\section{Our Approach}

We start by introducing our mathematical notation. Let $\mathcal{X}=\mathbb{R}^d$ be a $d$-dimensional input feature space, $L=\{l_1,l_2,...,l_q\}$ a label set containing $q$ labels and $\mathcal{Y}=\{0,1\}^q$ a $q$-dimensional label space. $D=\{(\bm{x}_i,\bm{y}_i)| 1 \leqslant i \leqslant n \}$ is a multi-label training data set containing $n$ instances. Each instance $(\bm{x}_i,\bm{y}_i)$ consists of a feature vector $\bm{x}_i \in \mathcal{X}$ and a label vector $\bm{y}_i \in \mathcal{Y}$, where $y_{ij}$ is the $j$-th element of $\bm{y}_i$ and $y_{ij}=1 (0)$ denotes that $l_j$ is (not) associated with $i$-th instance. 
A multi-label method learns the mapping function $h:\mathcal{X} \to \{0,1\}^q$ and (or) $f: \mathcal{X} \to \mathbb{R}^q$ from $D$ that given an unseen instance $\bm{x}$, outputs a label vector $\hat{\bm{y}}$ with the predicted labels of and (or) real-valued vector $\hat{\bm{fy}}$ corresponding relevance degrees to $\bm{x}$ respectively.

We propose a novel Multi-Label Synthetic Oversampling approach based on the Local distribution of labels (MLSOL). 
The pseudo-code of MLSOL is shown in Algorithm \ref{al:MLSOL}. Firstly, some auxiliary variables, as the weight vector $\bm{w}$ and type matrix $\bm{T}$ used for seed instance selection and synthetic examples generation respectively, are calculated based on the local label distribution of instances (line 3-6 in Algorithm \ref{al:MLSOL}). Then in each iteration, the seed and reference instances are selected, upon which a synthetic example is generated and added into the dataset. The loop (line 7-12 in Algorithm \ref{al:MLSOL}) would terminate when expected number of new examples are created. The following subsections detail the definition of auxiliaries as well as strategies to pick seed instances and create synthetic examples.

\begin{algorithm}
 \SetKwData{Left}{left}\SetKwData{This}{this}\SetKwData{Up}{up}
 \SetKwFunction{InitTypes}{InitTypes}\SetKwFunction{GenerateInstance}{GenerateInstance}
 \SetKwInOut{Input}{input}\SetKwInOut{Output}{output}
 
 \Input{multi-label data set: $D$, percentage of instances to generated: $P$, number of nearest neighbour: $k$ }
 \Output{new data set $D'$}
 $ GenNum \leftarrow |D|*P$ \tcc*[r]{number of instances to generate}
 $D' \leftarrow D$ \;
 Find the $k$NN of each instance \;
 Calculate $\bm{C}$ according to Eq.\eqref{eq:Cij} \;
 Compute $\bm{w}$ according to Eq.\eqref{eq:wi2} \;
 $\bm{T} \leftarrow \InitTypes(\bm{C},k)$ \tcc*[r]{Initialize the type of instances}
\While{$GenNum>0$}{
    Select a seed instance $(\bm{x}_s,\bm{y}_s)$ from $D$ based on the $\bm{w}$\;
    Randomly choose a reference instance $(\bm{x}_r,\bm{y}_r)$ from $kNN(\bm{x}_s)$ \;
    $(\bm{x}_{c},\bm{y}_{c}) \leftarrow \GenerateInstance \left( (\bm{x}_s,\bm{y}_s), T_s, (\bm{x}_r,\bm{y}_r), T_r \right)$\;
    $D' \leftarrow D' \cup (\bm{x}_{c},\bm{y}_{c})$ \;
    $GenNum \leftarrow GenNum-1$ \;
}
\KwRet{$D'$} \;
 	
 \caption{MLSOL}
  \label{al:MLSOL}
\end{algorithm}

\subsection{Selection of Seed Instances}

We sample seed instances with replacement, with the probability of selection being proportional to the minority class values it is associated with, weighted by the difficulty of correctly classifying these values based on the proportion of opposite (majority) class values in the local neighborhood of the instance. 

For each instance $x_i$ we first retrieve its $k$ nearest neighbours, $kNN(x_i)$. Then for each label $l_j$ we compute the proportion of neighbours having opposite class with respect to the class of the instance and store the result in the matrix $\bm{C} \in \mathbb{R}^{n \times q}$ according to the following equation, where $\llbracket \pi \rrbracket$ is the indicator function that returns 1 if $\pi$ is true and 0 otherwise:  

\begin{equation}
C_{ij}=\frac{1}{k}\sum_{\bm{x}_m \in kNN(\bm{x}_i)} \llbracket y_{mj} \neq y_{ij}\rrbracket
\label{eq:Cij}
\end{equation}

The values in $\bm{C}$ range from 0 to 1, with values close to 0 (1) indicating a safe (hostile) neighborhood of similarly (oppositely) labelled examples. A value of $C_{ij}=1$ can further be viewed as a hint that $x_i$ is an outlier in this neighborhood with respect to $l_j$.

The next step is to aggregate the values in $\bm{C}$ per training example, $x_i$, in order to arrive at a single sampling weight, $w_i$, characterizing the difficulty in correctly predicting the {\em minority} class values of this example. A straightforward way to do this is to simply sum these values for the labels where the instance contains the minority class. Assuming for simplicity of presentation that the value 1 corresponds to the minority class, we arrive at this aggregation as follows:

\begin{equation}
w_{i}=\sum_{j=1}^{q}C_{ij}\llbracket y_{ij} = 1\rrbracket
\label{eq:wi1}
\end{equation}

There are two issues with this. The first one is that we have also taken into account the outliers. We will omit them by adding a second indicator function requesting $C_{ij}$ to be less than 1. The second issue is that this aggregation does not take into account the global level of class imbalance of each of the labels. The fewer the number of minority samples, the higher the difficulty of correctly classifying the corresponding minority class. In contrast, Equation \ref{eq:wi1} treats all labels equally. To resolve this issue, we can normalize the values of the non-outlier minority examples in $\bm{C}$ so that they sum to 1 per label, by dividing with the sum of the values of all non-outlier minority examples of that label. This will increase the relative importance of the weights of labels with fewer samples.  Addressing these two issues we arrive at the following proposed aggregation: 

\begin{equation}
w_{i}=\sum_{j=1}^{q}\frac{C_{ij}\llbracket y_{ij} = 1\rrbracket \llbracket C_{ij} < 1\rrbracket}{\sum_{i=1}^{n}C_{ij}\llbracket y_{ij} = 1\rrbracket \llbracket C_{ij} < 1\rrbracket}
\label{eq:wi2}
\end{equation}

\subsection{Synthetic Instance Generation}

The definition of the type of each instance-label pair is indispensable for the assignment of appropriate labels to the new instances that we shall create.
Inspired by \cite{napierala2016types}, we distinguish minority class instances into four types, namely safe ($SF$), borderline ($BD$), rare ($RR$) and outlier ($OT$), according to the proportion of neighbours from the same (minority) class:
\begin{itemize}
    \item $SF$ :  $0 \leqslant C_{ij} < 0.3$. The safe instance is located in the region overwhelmed by minority examples.
    \item $BD$ :  $0.3 \leqslant C_{ij} < 0.7$. The borderline instance is placed in the decision boundary between minority and majority classes.
    \item $RR$ :  $0.7 \leqslant C_{ij} < 1$, and only if the type of its neighbours from the minority class are $RR$ or $OT$. Otherwise there are some $SF$ or $BD$ examples in the proximity, which suggests that it could be rather a $BD$. The rare instance, accompanied with isolated pairs or triples of minority class examples, is located in the majority class area and distant from the decision boundary.  
    \item $OT$ :  $ C_{ij}=1$. The outlier is surrounded by majority examples.
\end{itemize}

For the sake of uniform representation, the type of majority class instance is defined as majority ($MJ$). Let $\bm{T} \in \{SF,BD,RR,OT,MJ\}^{n \times q}$ be the type matrix and $T_{ij}$ be the type of $y_{ij}$. 
The detailed steps of obtaining $\bm{T}$ are illustrated in Algorithm \ref{al:InitTypes}.

 \begin{algorithm}[ht]
 \SetKwData{Left}{left}\SetKwData{This}{this}\SetKwData{Up}{up}
 \SetKwInOut{Input}{input}\SetKwInOut{Output}{output}
 
 \Input{The matrix storing proportion of kNNs with opposite class for each instance and each label: $\bm{C}$, number of nearest neighbour: $k$ }
 \Output{types of instances $\bm{T}$}
 \For(\tcc*[h]{$n$ is the number of instances}){$i \leftarrow 1$ \KwTo $n$}{
      \For (\tcc*[h]{$q$ is the number of labels}) {$j\leftarrow 1$ \KwTo $q$}{ 
        \eIf{$y_{ij}=$ majority class}{
            $T_{ij} \leftarrow MJ$ \;
        }
        (\tcc*[h]{$y_{ij}$ is the minority class}){
            \lIf{$ C_{ij}<0.3 $}{$T_{ij} \leftarrow SA$ }
            \lElseIf{$ C_{ij}<0.7 $}{$T_{ij} \leftarrow BD$ }
            \lElseIf{$ C_{ij}<1 $}{$T_{ij} \leftarrow RR$ }
            \lElse{$T_{ij} \leftarrow OT$ }
        }
     }
 }
 \Repeat(\tcc*[h]{re-examine $RR$ type}){\text{no change in} $\bm{T}$ }{
   \For{$i \leftarrow 1$ \KwTo $n$}{
      \For {$j\leftarrow 1$ \KwTo $q$}{ 
        \If{$T_{ij} = RR$}{
            \ForEach{$\bm{x}_m$ in $kNN(\bm{x}_i)$}{
                \If{$T_{mj}=SF$ \text{or} $T_{mj}=BD$ }{
                    $T_{ij} \leftarrow BD $\;
                    break \;
                }
            }
        }
      }
    } 
}
\KwRet{$\bm{T}$} \;
 \caption{InitTypes}
  \label{al:InitTypes}
\end{algorithm}

Once the seed instance $(\bm{x}_s,\bm{y}_s)$ has been decided, the reference instance $(\bm{x}_r,\bm{y}_r)$ is randomly chosen from the $k$ nearest neighbours of the seed instance. Using the selected seed and reference instance, a new synthetic instance is generated according to Algorithm \ref{al:GenerateInstance}.
The feature values of the synthetic instance $(\bm{x}_c,\bm{y}_c)$ are interpolated along the line which connects the two input samples (line 1-2 in Algorithm \ref{al:GenerateInstance}). Once $\bm{x}_c$ is confirmed, we compute $cd \in [0,1]$, which indicates whether the synthetic instance is closer to the seed ($cd < 0.5$) or closer to the reference instance ($cd > 0.5$) (line 3-4 in Algorithm \ref{al:GenerateInstance}).

\begin{algorithm}[ht]
 \SetKwData{Left}{left}\SetKwData{This}{this}\SetKwData{Up}{up}
 \SetKwFunction{Random}{Random}
 \SetKwInOut{Input}{input}\SetKwInOut{Output}{output}
 \Input{seed instance: $(\bm{x}_s,\bm{y}_s)$, types of seed instance: $T_s$, reference instance: $(\bm{x}_r,\bm{y}_r)$, types of reference instance: $T_r$}
 \Output{synthetic instance: $(\bm{x}_c,\bm{y}_c)$}
\For{$j\leftarrow 1$ \KwTo $d$} 
{ 
    $x_{cj} \leftarrow  x_{sj} +\Random(0,1)*(x_{rj}-x_{sj})$ \tcc*[r]{Random(0,1) generates a random value between 0 and 1}
}
$d_s \leftarrow dist(x_c,x_s)$, $d_r \leftarrow dist(x_c,x_r)$ \tcc*[r]{$dist$ return the distance between 2 instances}
$cd \leftarrow d_s/(d_s+d_r)$ \;
\For {$j\leftarrow 1$ \KwTo $q$}{ 
    \eIf{ $y_{sj}=y_{rj}$ }{
    $y_{cj} \leftarrow y_{sj}$ \;
    }{
        \If(\tcc*[h]{ensure $y_{sj}$ being minority class}){$T_{sj} = MJ$}{
            $s \longleftrightarrow r$ \tcc*[r]{swap indices of seed and reference instance}
            $cd \leftarrow 1-cd$ \;
        }
        \Switch{$T_{sj}$}{
            \lCase {$SF$}{ 
                $\theta \leftarrow 0.5$ ;  break 
            }
            \lCase {$BD$}{ 
                $\theta \leftarrow 0.75$ ;  break
            }\lCase {$RR$}{ 
                $\theta \leftarrow 1+1e-5$ ;  break 
            }\lCase {$OL$}{ 
                $\theta \leftarrow 0-1e-5$ ;  break
            }
        }
        \eIf{$cd \leqslant \theta$}{
            $y_{cj} \leftarrow y_{sj}$ \;    
        }{
            $y_{cj} \leftarrow y_{rj}$ \;   
        }
    }
   
}
\KwRet{$(\bm{x}_t,\bm{y}_t)$} \;
 \caption{GenerateInstance}
  \label{al:GenerateInstance}
\end{algorithm}

With respect to label assignment, we employ a scheme considering the labels and types of the seed and reference instances as well as the location of the synthetic instance, which is able to create informative instances for difficult minority class labels without bringing in noise for majority labels. 
For each label $l_j$, $y_{cj}$ is set as $y_{sj}$ (line 6-7 in Algorithm \ref{al:GenerateInstance}) if $y_{sj}$ and $y_{rj}$ belong to the same class.
In the case where $y_{sj}$ is majority class, the seed instance and the reference example should be exchanged to guarantee that $y_{sj}$ is always the minority class (line 9-11 in Algorithm \ref{al:GenerateInstance}). Then, $\theta$, a threshold for $cd$ is specified based on the type of the seed label, $T_{sj}$ (line 12-16 in Algorithm \ref{al:GenerateInstance}), which is used to determine the instance (seed or reference) whose labels will be copied to the synthetic example. 
For $SF$, $BD$ and $RR$, where the minority (seed) example is surrounded by several majority instances and suffers more risk to be classified wrongly, the cut-point of label assignment is closer to the majority (reference) instance. Specifically, $\theta=0.5$ for $SF$ represents that the frontier of label assignment is in the midpoint between seed and reference instance, $\theta=0.75$ for $BD$ denotes that the range of minority class extends as three times as large than the majority class, and $\theta >1 \geqslant cd$ for $RR$ ensures that the generated instance is always set as minority class regardless of its location. With respect to $OT$ as a singular point placed at majority class region, all possible synthetic instances are assigned the majority class due to the inability of an outlier to cover the input space.
Finally, $y_{cj}$ is set as $y_{sj}$ if $cd$ is not larger than $\theta$, otherwise $y_{cj}$ is equal to $y_{rj}$ (line 17-20 in Algorithm \ref{al:GenerateInstance}).

\begin{figure}
\begin{center}
\includegraphics[width=1.0\textwidth]{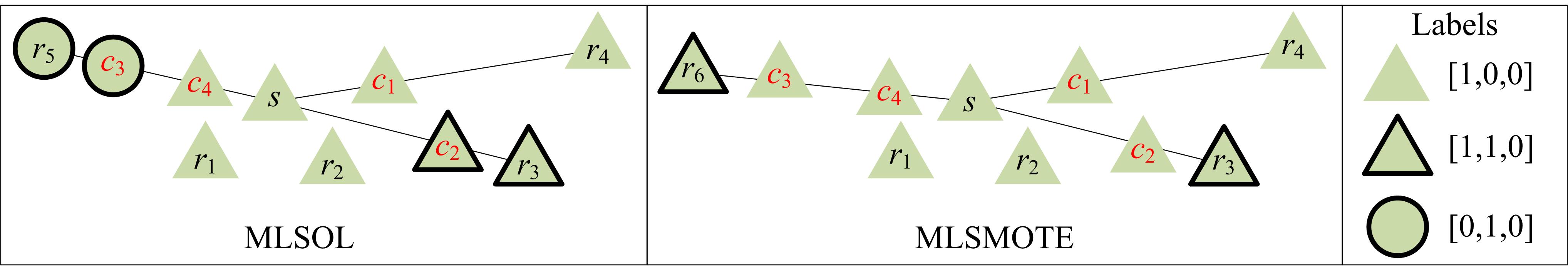}
\caption{An example of MLSOL excelling MLSMOTE. $s$ is the seed instance, $r_*$ are candidate reference instances ($k$NNs of $s$), and $c_*$ are possible synthetic examples. The synthetic instances created by MLSMOTE are associated with unique label vector ([1,0,0]) decided by predominant $k$NNs, while MLSOL assigns labels to new examples according to its location. The two sampling approaches are identical if the synthetic instance ($c_1$ and $c_4$) is near the instance whose labels are same with majority $k$NNs or seed instance, otherwise ($c_2$ and $c_3$) the MLSMOTE introduces noise while MLSOL could tackle it by copying the labels of nearest instance to the new example.
}
\label{fig:1}
\end{center}
\end{figure}

Compared with MLSMOTE, MLSOL is able to generate more diverse and well-labeled synthetic instances. As the example in Figure \ref{fig:1} shows, given a seed instance, the labels of the synthetic instance are fixed in MLSMOTE, while the labels of the new instance change according to its location in MLSOL, which avoids the introduction of noise as well. 

\subsection{Ensemble of Multi-Label Sampling (EMLS)}

Ensemble is a effective strategy to increase overall accuracy and overcome over-fitting problem, but has not been leveraged to multi-label sampling approaches. 
To improve the robustness of MLSOL and current multi-label sampling methods, we propose the ensemble framework called EMLS where any multi-label sampling approach and classifier could be embedded. In EMLS, $M$ multi-label learning models are trained and each model is built upon a re-sampled dataset generated by a multi-label sampling method with various random seed.
There are many random operations in existing and proposed multi-label learning sampling methods~\cite{Charte2015,Charte2015a}, which guarantees the diversity of training set for each model in the ensemble framework via employing different random seed. 
Then the bipartition threshold of each label is decided by maximizing F-measure on training set, as COCOA~\cite{Zhang2015a} and ECCRU3~\cite{Liu2018MakingImbalance} do. 
Given the test example, the predicting relevant scores is calculated as the average output relevant degrees obtained from $M$ models, and the labels whose relevance degree is larger than the corresponding bipartition threshold are predicted as "1", and "0" otherwise.    

\subsection{Complexity Analysis}
The complexity of searching $k$NN of input instances is $O(n^2d+n^2k)$. The complexity of computing $C$, $\bm{w}$ and $T$ is $O(knq)$, $O(nq)$ and $O(nq)$, respectively. The complexity of creating instances is $O(nP(n+d))$ where $nP$ is the number of generated examples. The overall complexity of MLSOL is $O(n^2d+n^2k+nkq)$, of which the $k$NN searching is the most time-consuming part.  

Let's define $\Theta_{t}(n,d,q)$ and $\Theta_{p}(d,q)$ the complexity of training and prediction of multi-label learning method respectively, and $\Theta_{s}(n,d,q)$ the complexity of a multi-label sampling approach. The complexity of EMLS is $O(M\Theta_{p}(d,q))$ for prediction and $O\left(M \left( \Theta_{s}(n,d,q)+\Theta_{t}(n,d,q)+n\Theta_{p}(d,q) \right) \right)$ for training.

\section{Empirical Analysis}

\subsection{Setup}

Table \ref{ta:Dataset} shows detailed information for the 13 benchmark multi-label datasets, obtained from Mulan's repository\footnote{\href{http://mulan.sourceforge.net/datasets-mlc.html}{http://mulan.sourceforge.net/datasets-mlc.html}}, that are used in this study. 
Besides, in textual data sets with more than 1000 features we applied a simple dimensionality reduction approach that retains the top 10$\%$ (bibtex, enron, medical) or top 1$\%$ (rcv1subset1, rcv1subset2, yahoo-Arts1, yahoo-Business1) of the features ordered by number of non-zero values (i.e. frequency of appearance). Besides, we remove labels only containing one minority class instance, because when splitting the dataset into training and test sets, there may be only majority class instances of those extremely imbalanced labels in training set.

\begin{table}[htp]
\centering
\small
\caption{The 16 multi-label datasets used in this study. Columns $n$, $d$, $q$ denote the number of instances, features and labels respectively, $LC$ the label cardinality, $MeanImR$ the average imbalance ratio of labels, where imbalance ratio of a label is computed as the number of majority instances divided by the number of minority instance of the label. 
}
\label{ta:Dataset}
\begin{tabular}{@{}ccccccc@{}}
\toprule
Dataset & Domain & $n$ & $d$ & $q$ & $LC$ & $MeanImR$ \\ \midrule
bibtex & text & 7395 & 183 & 159 & 2.402 & 87.7 \\
cal500 & music & 502 & 68 & 174 & 26 & 22.3 \\
corel5k & image & 5000 & 499 & 347 & 3.517 & 522 \\
enron & text & 1702 & 100 & 52 & 3.378 & 107 \\
flags & image & 194 & 19 & 7 & 3.392 & 2.753 \\
genbase & biology & 662 & 1186 & 24 & 1.248 & 78.8 \\
medical & text & 978 & 144 & 35 & 1.245 & 143 \\
rcv1subset1 & text & 6000 & 472 & 101 & 2.88 & 236 \\
rcv1subset2 & text & 6000 & 472 & 101 & 2.634 & 191 \\
scene & image & 2407 & 294 & 6 & 1.074 & 4.662 \\
yahoo-Arts1 & text & 7484 & 231 & 25 & 1.654 & 101 \\
yahoo-Business1 & text & 11214 & 219 & 28 & 1.599 & 286 \\
yeast & biology & 2417 & 103 & 14 & 4.237 & 8.954 \\ \bottomrule
\end{tabular}
\end{table}

Four multi-label sampling methods are used for comparison, namely the state-of-the-art MLSMOTE~\cite{Charte2015a} and RHwRSMT~\cite{Charte2019a} that integrates REMEDIAL~\cite{Charte2019} and MLSMOTE, as well as their ensemble versions, called EMLSMOTE and ERHwRSMT respectively. 
Furthermore, the base learning approach without employing any sampling approach, denoted as Default, is also used for comparing.
For all sampling methods, the number of nearest neighbours is set to 5 and the Euclidean distance is used to measure the distance between the examples. In MLSOL, the sampling ratio is set to 0.3. In RHwRSMT, the threshold for decoupling instance is set to $SCUMBLE$. For MLSMOTE and RHwRSMT, the label generation strategy is $Ranking$. The ensemble size is set to 5 for all ensemble methods. 
In addition, six multi-label learning methods are employed as base learning methods, comprising four standard multi-label learning methods (BR~\cite{Boutell2004}, MLkNN~\cite{zhang2007ml}, CLR~\cite{Furnkranz2008}, RAkEL~\cite{tsoumakas2011random}), as well as two state-of-the-art methods addressing the class imbalance problem (COCOA~\cite{Zhang2015a} and ECCRU3~\cite{Liu2018MakingImbalance}). 


Three widely used imbalance aware evaluation metrics are leveraged to measure the performance of methods, namely macro-averaged F-measure, macro-averaged AUC-ROC (area under the receiver operating characteristic curve) and macro-averaged AUCPR (area under the precision recall curve). For simplicity, we omit the ``macro-averaged" in further references to these metrics within the rest of this paper. 

The experiments were conducted on a machine with 4$\times$10-core CPUs running at 2.27 GHz. We apply $5 \times 2$-fold cross validation with multi-label stratification~\cite{Sechidis2011} to each dataset and the average results are reported. The implementation of our approach and the scripts of our experiments are publicly available at Mulan's GitHub repository\footnote{\href{https://github.com/tsoumakas/mulan/tree/master/mulan}{https://github.com/tsoumakas/mulan/tree/master/mulan}}. The default parameters are used for base learners.

\subsection{Results and Analysis}
Detailed experimental results are listed in the supplementary material of this paper. The statistical significance of the differences among the methods participating in our empirical study is examined by employing the Friedman test, followed by the Wilcoxon signed rank test with Bergman-Hommel's correction at the 5\% level, following literature guidelines~\cite{Garcia2008,Benavoli2016}. Table \ref{ta:Results} shows the average rank of each method as well as its significant wins/losses versus each one of the rest of the methods for each of the three evaluation metrics and each of the six base multi-label methods. The best results are highlighted with bold typeface. 

\begin{table}
\centering
\small
\caption{Average rank of the compared methods using 6 base learners in terms of three evaluation metrics. $A_1$, $A_2$, E$A_1$ and, E$A_2$ stands for MLSMOTE, RHwRSMT, EMLSMOTE and ERHwRSMT, respectively. The parenthesis ($n_1$/$n_2$) indicates the corresponding method is significantly superior to $n_1$ methods and inferior to $n_2$ methods based on the Wilcoxon signed rank test with Bergman-Hommel's correction at the 5\% level.}
\label{ta:Results}
\begin{tabular}{@{}cccccccc@{}}
\toprule
\multicolumn{8}{c}{F-measure} \\ \midrule
Base Method & Default & $A_1$ & $A_2$ & MLSOL & E$A_1$ & E$A_2$ & EMLSOL \\
BR & 5.19(1/4) & 3.73(2/2) & 7.00(0/6) & 4.23(2/2) & 2.08(5/1) & 4.38(1/2) & \textbf{1.38(6/0)} \\
MLkNN & 5.81(1/5) & 4.92(2/4) & 7.00(0/6) & 4.27(3/3) & \textbf{1.62(5/0)} & 2.69(4/2) & 1.69\textbf{(5/0)} \\
CLR & 5.04(1/3) & 4.5(1/3) & 7.00(0/6) & 4.15(1/3) & 2.58\textbf{(4/0)} & 2.62\textbf{(4/0)} & \textbf{2.12(4/0)} \\
RAkEL & 5.04(1/4) & 3.88(2/2) & 7.00(0/6) & 3.5(2/2) & 2.46(5/1) & 4.69(1/2) & \textbf{1.42(6/0)} \\
COCOA & 3.58(1/0) & 4.42(1/1) & 6.35(0/5) & 5.23(0/1) & 3.27(1/0) & 3.31(1/0) & \textbf{1.85(3/0)} \\
ECCRU3 & 3(2/0) & 4.58(1/1) & 6.31(0/5) & 5.46(0/2) & 3.35(1/0) & 3.46(1/1) & \textbf{1.85(4/0)} \\ \midrule
Total & 7/16 & 9/13 & 0/34 & 8/13 & 21/2 & 12/7 & \textbf{28/0} \\ \midrule
\multicolumn{8}{c}{AUC-ROC} \\ \midrule
Base Method & Default & $A_1$ & $A_2$ & MLSOL & E$A_1$ & E$A_2$ & EMLSOL \\
BR & 5.23(1/3) & 4.65(1/3) & 6.27(0/6) & 3.46(3/1) & 2.81(4/1) & 4.58(1/2) & \textbf{1.00(6/0)} \\
MLkNN & 4.69(1/1) & 3.73(2/2) & 6.35(0/6) & 4.23(2/1) & 2.58(3/1) & 5.35(1/4) & \textbf{1.08(6/0)} \\
CLR & 4.35(0/1) & 4.77(0/2) & 5.58(0/3) & 5.00(0/2) & 2.85(4/1) & 4.08(1/2) & \textbf{1.38(6/0)} \\
RAkEL & 4.38(2/4) & 3.73(3/2) & 6.77(0/6) & 3.54(3/2) & 2.54(5/1) & 6.00(1/5) & \textbf{1.04(6/0)} \\
COCOA & 5.23(0/1) & 4.73(0/1) & 5.42(0/1) & 4.54(0/1) & 3.42(0/1) & 3.65(0/1) & \textbf{1.00(6/0)} \\
ECCRU3 & 4.73(0/1) & 4.23(0/2) & 5.73(0/3) & 5.46(0/1) & 2.65(3/1) & 4.12(1/2) & \textbf{1.08(6/0)} \\ \midrule
Total & 4/11 & 6/12 & 0/25 & 8/8 & 19/6 & 5/16 & \textbf{36/0} \\ \midrule
\multicolumn{8}{c}{AUCPR} \\ \midrule
Base Method & Default & $A_1$ & $A_2$ & MLSOL & E$A_1$ & E$A_2$ & EMLSOL \\
BR & 4.81(1/2) & 3.85(1/2) & 6.46(0/6) & 4.15(1/2) & 2.46(5/1) & 5.19(1/2) & \textbf{1.08(6/0)} \\
MLkNN & 5.04(0/2) & 4.5(1/2) & 5.92(0/5) & 4.08(1/1) & 3.04(4/1) & 4.42(1/2) & \textbf{1.00(6/0)} \\
CLR & 4.15(1/1) & 4.88(0/2) & 5.92(0/4) & 5.00(0/3) & 3.00(3/1) & 3.81(2/1) & \textbf{1.23(6/0)} \\
RAkEL & 4.42(1/2) & 3.92(1/2) & 6.77(0/6) & 3.92(1/2) & 2.5(5/1) & 5.46(1/2) & \textbf{1.00(6/0)} \\
COCOA & 5.31(0/1) & 4.85(0/1) & 5.31(0/1) & 4.62(0/1) & 3.15(0/1) & 3.77(0/1) & \textbf{1.00(6/0)} \\
ECCRU3 & 4.96(0/2) & 4.5(0/2) & 5.50(0/3) & 5.04(0/2) & 2.88(5/1) & 4.12(1/2) & \textbf{1.00(6/0)} \\ \midrule
Total & 3/10 & 3/11 & 0/25 & 3/11 & 22/6 & 6/10 & \textbf{36/0} \\ \bottomrule

\end{tabular}
\end{table}

We start our discussion by looking at the single model version of the three resampling methods. We first notice that RHwRSMT achieves the worst results and that it is even worse than no resampling at all (default), which is mainly due to the additional bewilderment yielded by REMEDIAL, i.e. there are several pairs of instances with same features and disparate labels. MLSOL and MLSMOTE exhibit similar total wins and losses, especially in AUCPR, which is considered as the most appropriate measure in the context of class imbalance \cite{Saito2015}. Moreover, the wins and losses of MLSOL and MLSMOTE are not that different from no resampling at all. This is particularly true when using a multi-label learning method that already handles class imbalance, such as COCOA and ECCRU3, which is not surprising. 

We then  notice that the ensemble versions of the three multi-label resampling methods outperform their corresponding single model versions in all cases. This verifies the known effectiveness of resampling apporaches in reducing the error, in particular via reducing the variance component of the expected error \cite{Hastie2016}. Ensembling enables MLSMOTE and MLSOL to achieve much better results compared to no resampling and it even helps RHwRSMT to do slightly better than no resampling.  

Focusing on the ensemble versions of the three resampling methods we notice that EMLSOL achieves the best average rank and the most significant wins without suffering any significant loss in all 18 different pairs of the 6 base multi-label methods and the 3 evaluation measures, with the exception that MLSMOTE with MLkNN as base learner achieves best average rank in terms of F-measure. EMLSMOTE comes second in total wins and losses in most cases, while ERHwRSMT does much worse than EMLSMOTE. 

An interesting observation here is that while MLSOL and MLSMOTE have similar performance, MLSOL benefitted much more than MLSMOTE from the ensemble approach. This happens because randomization plays a more important role in MLSOL than in MLSMOTE. MLSOL uses weighted sampling for seed instance selection, while MLSMOTE takes all minority samples into account instead. This allows EMLSOL to create more diverse models, which achieve greater error correction when aggregated.

\section{Conclusion}
We proposed MLSOL, a new synthetic oversampling approach for tackling the class-imbalance problem in multi-label data. Based on the local distribution of labels, MLSOL selects more important and informative seed instances and generates more diverse and well-labeled synthetic instances. In addition, we employed MLSOL within a simple ensemble framework, which exploits the random aspects of our approach during sampling training examples to use as seeds and during the generation of synthetic training examples.  


We experimentally compared the proposed approach against two state-of-the art resampling methods on 13 benchmark multi-label datasets. The results offer strong evidence on the superiority of MLSOL, especially of its ensemble version, in three different imbalance-aware evaluation measures using six different underlying base multi-label methods.


\subsubsection{Acknowledgements} Bin Liu is supported from the China Scholarship Council (CSC) under the Grant CSC No.201708500095.

%
%
%

\bibliographystyle{splncs04} %


\end{document}